\documentclass{article} % For LaTeX2e
\usepackage{iclr2023_conference_tinypaper,times}

\usepackage{amsmath,amsfonts,bm}

\def\eqref#1{equation~\ref{#1}}
\def\1{\bm{1}}

\def\vx{{\bm{x}}}

\DeclareMathAlphabet{\mathsfit}{\encodingdefault}{\sfdefault}{m}{sl}
\SetMathAlphabet{\mathsfit}{bold}{\encodingdefault}{\sfdefault}{bx}{n}

\usepackage{makecell} % Xhline
\usepackage{multirow} % Table 기능
\usepackage{adjustbox} % \resizebox{\textwidth}{!}
\usepackage{hyperref}
\usepackage{url}

\title{Adverb Is the Key: Simple Text Data \\ Augmentation with Adverb Deletion}

\author{Juhwan Choi and Youngbin Kim \\
Chung-Ang University, Seoul, Republic of Korea \\
\texttt{\{gold5230, ybkim85\}@cau.ac.kr} \\
}

\iclrfinalcopy % Uncomment for camera-ready version, but NOT for submission.
\begin{document}
\maketitle

\begin{abstract}
In the field of text data augmentation, rule-based methods are widely adopted for real-world applications owing to their cost-efficiency. However, conventional rule-based approaches suffer from the possibility of losing the original semantics of the given text. We propose a novel text data augmentation strategy that avoids such phenomena through a straightforward deletion of adverbs, which play a subsidiary role in the sentence. Our comprehensive experiments demonstrate the efficiency and effectiveness of our proposed approach for not just single text classification, but also natural language inference that requires semantic preservation. We publicly released our source code for reproducibility.
\end{abstract}

\section{Introduction}

Text data augmentation is an important regularization technique to alleviate overfitting and improve the robustness of an NLP model. Even though various strategies have been proposed, text data augmentation methods often involve a trade-off between complexity and potential performance gain \citep{feng2021survey}. While rule-based methods \citep{zhang2015character, belinkov2018synthetic, wei-zou-2019-eda, karimi-etal-2021-aeda-easier, choi2023softeda} are simple and easy to implement, they often introduce small diversity, which suffers to enhance performance significantly \citep{zhang2022good}. Whereas, relatively complex methods that utilize deep learning models \citep{sennrich-etal-2016-improving, wu2019conditional, anaby2020not} may acquire different expressions, which leads to potentially large performance improvement. However, deploying them for real-world applications can be expensive, as they often necessitate additional deep learning models. This becomes particularly restrictive when hardware resources are limited, as is often the case for projects in developing countries.

Consequently, rule-based approaches are widely employed for real-world applications owing to their low cost. However, unlike rule-based image augmentation such as flipping or cropping, which maintains the original contents, rule-based text data augmentation methods have another challenge of semantic preservation. The conventional approach of rule-based text augmentation methods relies on introducing a perturbation through predefined rules, such as synonym replacement, random insertion, random swap, and random deletion \citep{wei-zou-2019-eda}. While previous researchers have recognized this drawback and proposed several refinements, they tend to introduce less variation \citep{karimi-etal-2021-aeda-easier} or not fully maintain the semantics of the original sentence \citep{choi2023softeda}.

In light of these previous studies, we propose a straightforward yet novel rule-based text augmentation method by deleting adverbs from the given dataset. Adverbs mostly focus on adjusting another word, by maximizing or diminishing the meaning of other words \citep{delfitto2006adverb, ruppenhofer2015ordering}. Through the explicit removal of these adverbs and keeping more important words such as nouns and verbs, we can attain new sentences while maintaining the core semantics of the original sentence.

To the best of our knowledge, this work is the first study to actively focus on the role of adverbs, which has often taken less interest \citep{nikolaev2023adverbs}, for text data augmentation. We evaluated our approach on various text classification tasks, as well as natural language inference (NLI) tasks, which are more complex compared to single text classification. Our empirical experimental result showed the effectiveness and efficiency of the proposed method, especially on NLI tasks, which require further semantic preservation compared to single text classification.

\section{Method}

We aim to synthesize the given data $\vx=\{w_1, w_2, ..., w_n\}$ that comprises the dataset $\mathcal{D}$ and generate $\hat{\vx}$, the modified version of $\vx$ following the predefined rule. Previous methods such as easy data augmentation (EDA) and an easier data augmentation (AEDA) have applied this rule on a random word, regardless of the importance of each word. However, the proposed method aims to remove only adverbs from the given sentence. This process can be formulated as follows:

\begin{align*}
  \hat{\vx} = \vx \setminus \{W_{adv}\},\;where\;\{W_{adv}\} = \texttt{POS\_Tagger}(\vx, \text{ADV})
\end{align*}

In this formulation, $\{W_{adv}\}$ denotes a set of words that are tagged as adverbs by a part-of-speech (POS) tagger. We skip $\vx$ that does not have at least one adverb from the augmentation process.

\section{Experiment}

\begin{table}[h]
\caption{Accuracy (\%) and performance gain (\%p) across eight datasets. The best values for each dataset are boldfaced. Results that reported a lower value than the baseline are colored in gray.}
\label{tab-main}
\begin{center}
\resizebox{\textwidth}{!}{
\begin{tabular}{l|cccccccc}
\Xhline{3\arrayrulewidth}
\multicolumn{1}{c|}{\textbf{BERT}}   & SST2           & SST5           & CoLA                    & TREC           & RTE                     & MNLI-M                  & MNLI-MM                   & QNLI \\ \Xhline{2\arrayrulewidth}
No Aug                               & 88.74          & 50.58          & 79.24                   & 95.08          & 64.72                   & 75.35                   & 77.17                     & 85.81\\
EDA                                  & 89.18          & 50.36          & \textcolor{gray}{76.80} & 95.27          & \textcolor{gray}{62.43} & \textcolor{gray}{74.51} & \textcolor{gray}{75.85}   & \textcolor{gray}{83.17}  \\
AEDA                                 & 89.41          & 50.63          & 79.45                   & 95.39          & 65.92                   & \textcolor{gray}{74.60} & 77.35                     & 86.55   \\
softEDA                              & 89.24          & 50.89          & \textcolor{gray}{76.32} & 95.86          & 65.41                   & \textcolor{gray}{74.06} & \textcolor{gray}{75.93}   & \textcolor{gray}{84.97}    \\
Ours                                 & \textbf{89.73} & 51.25          & 80.28                   & 96.48          & 65.58                   & 76.11                   & 77.68                     & 87.07    \\ 
Ours w/ Curr.                        & 89.52          & \textbf{52.37} & \textbf{83.15}          & \textbf{96.95} & \textbf{68.72}          & \textbf{77.97}          & \textbf{78.14}            & \textbf{87.21}    \\ \Xhline{3\arrayrulewidth}
\end{tabular}
}
\end{center}
\end{table}

To evaluate the effectiveness of the proposed method, we conducted the experiment on various text classification and NLI datasets, in comparison with previous rule-based text augmentation methods. We built each model using BERT \citep{devlin2019bert} as the baseline model. For comparison with previous methods, we adopted EDA \citep{wei-zou-2019-eda}, AEDA \citep{karimi-etal-2021-aeda-easier}, and softEDA \citep{choi2023softeda}, a method that compensates semantic damage through soft label. More details about implementation details and datasets used for experiment can be found in Appendix~\ref{app-implementation-details} and \ref{app-dataset-specs}.

Table~\ref{tab-main} demonstrates the experimental result. The results indicate that the proposed method is able to enhance the performance of each model in various tasks, including NLI, where previous methods mostly suffered from performance degradation. This is due to the complexity of NLI tasks, which require a proper connection between two sentences after the augmentation, prohibiting semantic damage after the augmentation. These results clearly showcase the superiority of the proposed method in terms of preventing semantic damage while providing simple and low-cost augmentation technique. Moreover, our additional experiment that combines our method with curriculum data augmentation \citep{wei2021few, ye2021efficient, lu2023pcc} reveals further performance gain.

\section{Conclusion}

We proposed a novel rule-based text data augmentation method focusing on adverbs. Through this straightforward approach, it is able to easily preserve the original meaning of the given sentence compared to previous rule-based text augmentation methods. This advantage led to performance gain on not just text classification tasks but also NLI tasks, where other rule-based methods suffer to enhance the performance. Future research could expand our method to other tasks, including text summarization. Additionally, we are planning to extend this approach to other languages, based on the universality of adverbs \citep{delfitto2006adverb}.

\subsubsection*{Acknowledgements}
This research was supported by Basic Science Research Program through the National Research Foundation of Korea(NRF) funded by the Ministry of Education(NRF-2022R1C1C1008534), and Institute for Information \& communications Technology Planning \& Evaluation (IITP) through the Korea government (MSIT) under Grant No. 2021-0-01341 (Artificial Intelligence Graduate School Program, Chung-Ang University).

\subsubsection*{URM Statement}
First author Juhwan Choi meets the URM criteria of ICLR 2024 Tiny Papers Track. He is outside the range of 30-50 years, non-white researcher.

\bibliography{iclr2023_conference_tinypaper}
\bibliographystyle{iclr2023_conference_tinypaper}

\appendix

\section{Implementation Details}
\label{app-implementation-details}
This section describes experimental setups and implementation details for reproduction. We have mainly followed the source code provided by softEDA \citep{choi2023softeda}. Please refer to our source code for further investigation.\footnote{\url{https://github.com/c-juhwan/adverb-deletion-aug}}

\textbf{Augmentation Technique.} We used a POS tagger offered by spaCy \citep{honnibal2020spacy} library, with \texttt{en\_core\_web\_sm} model. We removed all words that are tagged as \texttt{ADV} from each sentence of a dataset. We skipped a sentence with no adverb from the augmentation process. For softEDA baseline, we used $\alpha=0.2$ for every experiment.

For curriculum data augmentation, we utilized two-stage curriculum learning suggested by previous researchers \citep{wei2021few}. Specifically, we first trained the initial model using only original data for 2 epochs and embraced augmented data for the rest 3 epochs of the training procedure.

\textbf{Model and Dataset.} The model and datasets were mainly implemented with Hugging Face Transformers \citep{wolf2020transformers} and Datasets \citep{lhoest2021datasets} libraries. The BERT model has used \texttt{bert-base-uncased}. We note that the NLI datasets used in our experiments are drawn from the GLUE \citep{wang2019glue} benchmark and the test set is not publicly available. Instead, we used the validation set as the test set and $N_{Test}$ denotes the amount of validation data. Similar to datasets that do not have a predefined validation set, we used randomly selected 20\% of the training set as the validation set for these datasets.

\textbf{Hyperparameters.} Adam \citep{kingma2015adam} has been deployed as the optimizer, with a learning rate of 1e-4. We trained each model for 5 epochs with a batch size of 32.

\textbf{Further Details.} In cases where no predefined validation set exists, we randomly selected 20\% of the training data as the validation set. The training procedure was performed with a single NVIDIA RTX 3090 GPU.

\section{Case Analysis and Discussion}
\label{app-case-analysis}

Table~\ref{tab-case-analysis} displays examples of augmentation results using the traditional EDA technique and our proposed method. The examples show that our method generates more syntactically acceptable, and semantically consistent sentences compared to EDA. This superiority comes from the randomness of EDA techniques, such as deleting important words or swapping the order within the sentence. Whereas, our proposed method could generate better samples through targeted deletion of adverbs, which play supplementary roles in the sentence. 

However, while our method has a distinct advantage, we acknowledge an existing limitation. As our method is solely composed of the deletion of adverbs, the scope of variation is relatively small. This can be interpreted as a trade-off between semantic consistency and diverse augmentation. Nonetheless, the variation is larger than that of AEDA, which only injects punctuation marks, as our method aims to delete several words from the original sentence. Furthermore, in a real-world scenario, the engineer could combine our method with previous methods to acquire further performance boost. Future work could extend our approach to ensure more variability of augmented samples.

\begin{table}[h]
\caption{Comparison between EDA and our proposed method on examples from the SST2 dataset.}
\label{tab-case-analysis}
\begin{center}
\resizebox{\textwidth}{!}{
\begin{tabular}{l|l|l}
\Xhline{3\arrayrulewidth}
\multicolumn{1}{c|}{Original}                                                                                                                                                          & \multicolumn{1}{c|}{EDA}                                                                                                                                                  & \multicolumn{1}{c}{Ours}                                                                                                                                                                   \\ \Xhline{2\arrayrulewidth}
The film is strictly routine.                                                                                                                                                          & The is strictly film routine.                                                                                                                                             & The film is routine.                                                                                                                                                   \\ \hline
\begin{tabular}[c]{@{}l@{}}The santa clause 2 proves itself\\ a more streamlined and thought \\ out encounter than the original\\ could ever have hoped to be.\end{tabular}            & \begin{tabular}[c]{@{}l@{}}The santa clause 2 itself a \\ streamlined and original \\ thought out encounter than \\ the could have to be.\end{tabular}                     & \begin{tabular}[c]{@{}l@{}}The santa clause 2 proves itself\\ a streamlined and thought out\\ encounter than the original\\ could have hoped to be.\end{tabular}       \\ \hline
\begin{tabular}[c]{@{}l@{}}This is a very ambitious project \\ fora fairly inexperienced \\ filmmaker, but good actors, \\ good poetry and good music \\ help sustain it.\end{tabular} & \begin{tabular}[c]{@{}l@{}}This is a very ambitious for \\ a fairly inexperienced \\ actors, but filmmaker good, \\ good poetryand good music \\ sustain it.\end{tabular} & \begin{tabular}[c]{@{}l@{}}This is a ambitious project for\\ a inexperienced filmmaker, \\ but good actors, good poetry\\ and good music help sustain it.\end{tabular} \\ \hline
\begin{tabular}[c]{@{}l@{}}Perhaps the best sports \\ movie i've ever seen.\end{tabular}                                                                                               & Best perhaps movie seen.                                                                                                                                                  & the best sports movie i've seen.                                                                                                                                       \\ \Xhline{3\arrayrulewidth}
\end{tabular}
}
\end{center}
\end{table}

\newpage
\section{Dataset Specifications}
\label{app-dataset-specs}

\begin{table}[h]
\caption{Dataset used for the experiment. The datasets with $*$ indicate that these datasets used the validation set as the test set, as stated in Appendix~\ref{app-implementation-details}.}
\label{tab-dataset-specs}
\begin{center}
\resizebox{\textwidth}{!}{
\begin{tabular}{l|cccc}
\Xhline{3\arrayrulewidth}
\multicolumn{1}{c|}{\textbf{Dataset}}                 & Task  & $N_{Class}$  & $N_{Train}$ & $N_{Test}$  \\ \Xhline{2\arrayrulewidth}
SST2 \citep{socher-etal-2013-recursive}               & Sentiment & 2 & 6,919 & 1,820 \\
SST5 \citep{socher-etal-2013-recursive}               & Sentiment & 5 & 8,544 & 2,210 \\
% CR \citep{hu2004mining} \citep{qian2015automated}   & Sentiment & 2 & 3,011 & 752   \\
% MR \citep{pang-etal-2002-thumbs}                    & Sentiment & 2 & 9,593 & 1,067 \\
TREC \citep{li-roth-2002-learning}                    & Question Type & 6 & 5,452 & 500 \\
% SUBJ \citep{pang-lee-2004-sentimental}              & Subjectivity & 2 & 8,000 & 2,000 \\
% PC \citep{ganapathibhotla-liu-2008-mining}          & Pro-Con & 2 & 39,418 & 4,506 \\
CoLA \citep{warstadt-etal-2019-neural}                & Acceptability & 2 & 8,551 & 527 \\
RTE$^{*}$ \citep{dagan2005pascal}                     & NLI & 2 & 2,490 & 277 \\
MNLI-matched$^{*}$ \citep{williams2018broad}          & NLI & 3 & 392,702 & 9,815 \\
MNLI-mismatched$^{*}$ \citep{williams2018broad}       & NLI & 3 & 392,702 & 9,832 \\
QNLI$^{*}$ \citep{rajpurkar2016squad}                 & NLI & 2 & 104,743 & 5,463 \\ \Xhline{3\arrayrulewidth}

\end{tabular}
}
\end{center}
\end{table}

\end{document}